\documentclass[lettersize,journal]{IEEEtran}
\usepackage{amsmath,amsfonts}
\usepackage{algorithm}
\usepackage{algorithmic}
\usepackage{array}
\usepackage[caption=false,font=normalsize,labelfont=sf,textfont=sf]{subfig}
\usepackage{textcomp}
\usepackage{stfloats}
\usepackage{url}
\usepackage{verbatim}
\usepackage{graphicx}
\usepackage{array,booktabs}
\usepackage{multirow}
\usepackage{threeparttable}
\usepackage[colorlinks, linkcolor=red, citecolor=green]{hyperref}
\def\BibTeX{{\rm B\kern-.05em{\sc i\kern-.025em b}\kern-.08em
    T\kern-.1667em\lower.7ex\hbox{E}\kern-.125emX}}
\usepackage{balance}
\begin{document}

\title{Dynamic High-frequency Convolution for Infrared Small Target Detection}
\author{
    Ruojing Li, Chao Xiao, Qian Yin, Wei An, Nuo Chen, Xinyi Ying, Miao Li, Yingqian Wang

\thanks{This paper was partially supported by the National Natural Science Foundations of China under Grants 62501609, 42501589, 62401590, and 62505362.}
\thanks{The authors are with the College of Electronic Science and Technology, National University of Defense Technology, Changsha 410073, China. Corresponding authors are Wei An and Miao Li.}}

\maketitle

\begin{abstract}
Infrared small targets are typically tiny and locally salient, which belong to high-frequency components (HFCs) in images. Single-frame infrared small target (SIRST) detection is challenging, since there are many HFCs along with targets, such as bright corners, broken clouds, and other clutters. Current learning-based methods rely on the powerful capabilities of deep networks, but neglect explicit modeling and discriminative representation learning of various HFCs, which is important to distinguish targets from other HFCs. To address the aforementioned issues, we propose a dynamic high-frequency convolution (DHiF) to translate the discriminative modeling process into the generation of a dynamic local filter bank. Especially, DHiF is sensitive to HFCs, owing to the dynamic parameters of its generated filters being symmetrically adjusted within a zero-centered range according to Fourier transformation properties. Combining with standard convolution operations, DHiF can adaptively and dynamically process different HFC regions and capture their distinctive grayscale variation characteristics for discriminative representation learning. DHiF functions as a drop-in replacement for standard convolution and can be used in arbitrary SIRST detection networks without significant decrease in computational efficiency. To validate the effectiveness of our DHiF, we conducted extensive experiments across different SIRST detection networks on real-scene datasets. Compared to other state-of-the-art convolution operations, DHiF exhibits superior detection performance with promising improvement. Codes are available at \href{https://github.com/TinaLRJ/DHiF}{https://github.com/TinaLRJ/DHiF}.
\end{abstract}

\section{Introduction}

Single-frame infrared small target (SIRST) detection \cite{zhang2022exploring, zhang2023dim2clear} has been a long-standing and essential problem, with wide applications in maritime surveillance \cite{ribeiro2017data, zhou2019background}, UAV monitoring \cite{huang2023anti, li2025probing}, traffic management \cite{zhang2022rkformer, zhang2024irprunedet}, and beyond \cite{li2023direction, guo2024dmfnet, li2024multi, zeng2024mmi}. SIRST detection has faced the fundamental challenge of isolating small, textureless targets from complex and cluttered infrared backgrounds. A critical factor leading to this challenge is that both the targets of interest and the primary sources of background interference manifest predominantly as high-frequency components (HFCs) in infrared images. HFCs arise from rapid spatial grayscale variations and can be categorized into two distinct types according to their semantics.

\begin{itemize}
\item Genuine small targets: The signals of interest, typically characterized by localized intensity peaks or specific patterns (e.g., blob-like, point-like) against the background.
\item Structural background clutters: The signals of no interest, originating from intricate scene structures, such as sharp edges, corners, and target-like fragments (e.g., isolated bright spots, broken clouds, or man-made small objects).
\end{itemize}

The core difficulty of SIRST detection is to distinguish targets from clutters, since they often occupy overlapping high frequencies and exhibit superficially similar local characteristics (e.g., high local contrast). However, their underlying semantic meanings and grayscale variation patterns are significantly different. Consequently, achieving high-performance detection necessitates explicitly modeling and differentiating different categories of HFCs.

While traditional SIRST detection methods \cite{wei2016multiscale, zhu2020balanced} exploit HFCs' properties (mainly by modeling targets and clutters) for detection, they typically rely on hand-made and fixed-parameter operators, e.g., local contrast measurement \cite{wei2016multiscale} and top-hat transformation \cite{zhu2020balanced}. These operators are designed based on prior assumptions about target or clutter characteristics, and struggle to dynamically and effectively differentiate different HFCs. Thus, they limit the generality and robustness of traditional methods. Recently, deep learning-based methods \cite{wu2023mtu, guo2025infrared} have emerged with powerful data-driven modeling capabilities, offering the potential to overcome the limitations of fixed operators. They have achieved significant performance gains in many fields \cite{zhao2025chirplet, wang2025hypersigma, liu2024spatial, liuRandomFeatures, chen2025event}. However, most existing deep learning-based methods focus primarily on free feature fitting from annotated data, while overlooking the explicit modeling and discriminative representation learning of the semantic categories of HFCs within local regions.  

To address these issues, we propose a new dynamic high-frequency convolution (DHiF) for SIRST detection to enable explicit discriminative modeling of different HFC categories. To ensure the sensitivity of DHiF to HFCs, the kernel parameters of generated filters are symmetrically adjusted within a zero-centered range, according to the differentiation and linearity properties of the Fourier transformation. Specifically, this discriminative modeling is performed through dynamically generating specialized filter operators conditioned on local input features. The adaptability of DHiF allows networks to learn discriminative representations of various HFCs. Designed as a drop-in replacement for standard convolution, DHiF can be seamlessly integrated into arbitrary SIRST detection frameworks with convolutional structures. Extensive experiments on different real-scene datasets demonstrate the generality and effectiveness of DHiF in SIRST detection.

The main contributions of this work are as follows.

\begin{itemize}
    \item We emphasize the importance of learning discriminative representations of different HFCs and propose DHiF, a dynamic high-frequency convolution operator that enables discriminative HFC modeling via dynamically generated filter banks.
    \item We design DHiF as a drop-in replacement for standard convolutions within the encoders of arbitrary SIRST detection networks for consistent performance enhancement across various targets and scenarios.
    \item We conducted extensive experiments with various detection frameworks. The results demonstrate that our DHiF enables discriminative representation learning and improves the predictive capability of SIRST detection networks with negligible changes to the computational efficiency.
\end{itemize}

\section{Methodology}
In this section, we first analyze the actually used features of various existing SIRST detection networks, demonstrating the importance of HFC information. Then, we elaborate on the derivation and realization details of our DHiF and its seamless replacement into current SIRST detection methods.

\subsection{Importance of HFC Representation Learning}
As noted, both targets and clutters belong to HFCs in the images. Theoretically, it is important to learn the representations of different HFCs. Traditional methods \cite{zhu2020balanced, yang2023small} with strong interpretability are precisely based on that and achieve detection by explicitly modeling HFCs. As for deep learning-based methods that rely on data mining to learn key information for prediction, is the representation learning of different HFCs the universal core? To get a definite reply, we employed the attribution analysis \cite{gu2021interpreting} to interpret the target predictions across various networks, e.g., DNANet \cite{li2022dense}, ILNet \cite{li2025ilnet}, and SCTransNet \cite{yuan2024sctransnet}. The visualized attribution maps are shown in Fig. \ref{attribution}. More visualizations of more methods are given in the supplementary material.

\begin{figure}[t]
\centering
\includegraphics[width=1\columnwidth]{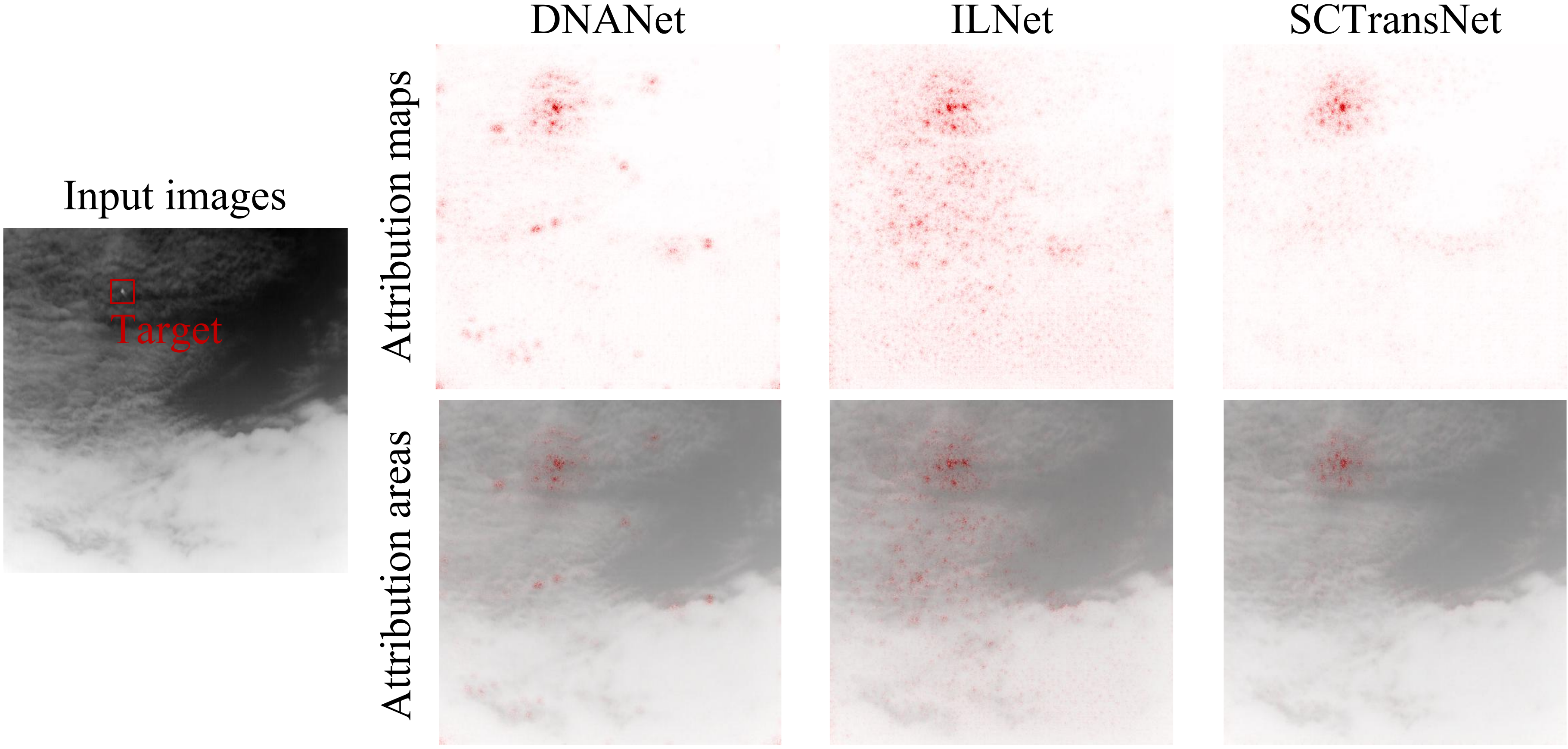}
\caption{Attribution visualizations of different methods for the predictions of target regions. The right three columns of subplots present the attribution maps generated by different methods, followed underneath by overlay visualizations with the input images. The red regions indicate contributions to the target predictions, with color intensity corresponding to the influence magnitude.}
\label{attribution}
\end{figure}

We can observe the key information captured by the networks through the attribution visualizations. Firstly, the most influential regions are located within the immediate neighborhoods of the targets, since the target information (e.g., local saliency) is essential for prediction. Besides target-centric regions, many background regions with textural patterns (e.g., cloud edges, building contours, vegetation textures) also significantly impact target predictions. This indicates that besides target information, the information of background HFCs is also important for detection.

These observations indicate that different methods focus on intensity gradient regions (especially high-frequency regions) and learn their representations for accurate prediction. Therefore, learning representations of HFCs is important for SIRST detection networks.

\subsection{Dynamic High-frequency Convolution}
To better distinguish targets from backgrounds, it is necessary to lead the networks to learn the discriminative representations of different HFCs (not only their individual representations). For this purpose, we explore how to dynamically model different HFCs within networks. The modeling results are usually refined into the used operators (e.g., the filters in traditional methods). The operators should be dynamically changing, since the HFCs in different regions are always different. Therefore, in this paper, we propose our dynamic high-frequency convolution (DHiF), which uses machine learning to automatically generate dynamic local filter bank. In this part, we introduce our DHiF in detail. Firstly, we formalize how to implement convolution operations with predefined filters, which is the basis of our method. Then, we analyze how to make DHiF sensitive to HFCs and describe the structure of DHiF. Finally, we analyze the drop-in replacement of DHiF.

\begin{figure}[t]
\centering
\includegraphics[width=0.7\columnwidth]{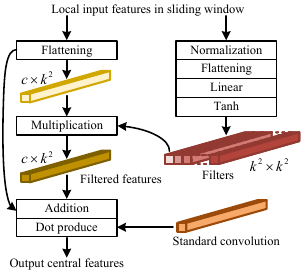}  
\caption{DHiF structure visualization. A dynamic local filter bank is generated via transformations based on local input features. The filters process local features to produce filtered high-frequency representations. Standard convolution then integrates filtered features with original input to yield output features.} 
\label{DHiF}
\end{figure}

\subsubsection{Convolutions with predefined filters}
Let an image $\boldsymbol{X}\in\mathbb{R}^{H\times W}$ and a sliding window of size $k\times k$ denote the input and the processing window of convolution. To capture the interesting property within a local region $\boldsymbol{x}\in\mathbb{R}^{k\times k}$, specific operations are executed in a sliding window for the local filtered feature. Note that if the targeted property is location-dependent within the window, a bank of filters becomes necessary for each pixel in the window to handle positional dependencies.

Therefore, the convolution and filtering operations share one sliding window. Firstly, $k \times k$ filters are applied to a local region, and the local filtered features are generated, i.e., $\boldsymbol{f}(\boldsymbol{x})\in \mathbb{R}^{k\times k}$. That is,
\begin{equation}
    \label{equ_fx}
    \boldsymbol{f}(\boldsymbol{x}) = [f_1(\boldsymbol{x}), f_2(\boldsymbol{x}), ..., f_{k\times k}(\boldsymbol{x})],
\end{equation}
where $f_i(\boldsymbol{x})$ is for the $i^{th}$ pixel in $\boldsymbol{x}$, i.e., $x_i$. By scanning the input image $\boldsymbol{X}$ with unit stride to perform filtering, the filtered features $\boldsymbol{f}(\boldsymbol{X}) \in \mathbb{R}^{H\times W\times k\times k}$ of all local regions are generated. Then, the convolution operation is performed on the filtered features, which can be described as
\begin{equation}
    \label{equ_F2}
    \boldsymbol{F} = \boldsymbol{f}(\boldsymbol{X}) \cdot \boldsymbol{W}.
\end{equation}
Finally, the output feature map $\boldsymbol{F}$ is generated.

\subsubsection{Sensitivity to HFCs}
The core of DHiF is to generate filter banks sensitive to HFCs. To implement DHiF, a prerequisite needs to be clarified, and a key issue needs to be resolved.

It is necessary to clarify what kind of filters are sensitive to HFCs. According to the differentiation property of the Fourier transformation, for a signal $f(t)$ with the Fourier transform $F(\omega)$, the operation $\frac{d}{dt}f(t)$ transforms to $j\omega F(\omega)$. That is,
\begin{equation}
    \label{equ_fx}
    f^{(n)}(t) \leftrightarrow (j\omega)^n F(\omega),
\end{equation}
where $t$ denotes time, and $\omega$ denotes frequency. $f^{(n)}(t)$ represents the $n^{th}$ derivative of $f(t)$. To enhance the responses of filters at high frequencies, spatial differentiation operations should be incorporated into filters. Simultaneously, according to the linearity property of the Fourier transform, adding a direct current signal ($g(t)=C$, $C \in \mathbb{R}$) in the time domain corresponds to adding an impulse function $\delta(\omega)$ at zero frequency in the frequency domain. That is,
\begin{equation}
    \label{equ_fx}
    f(t) + g(t) \leftrightarrow F(\omega) + C\cdot 2\pi \delta(\omega).
\end{equation}
To reduce the responses at low frequencies, constant kernels should be avoided from being included in filters. Therefore, the kernel parameters of a filter should contain both positive and negative values, and their sum should be close to zero.

Based on these analyses, in this paper, we address the issue of the sensitivity of the generated filters to HFC by restricting the kernel parameters of a filter to be within an interval centered at zero.

\subsubsection{Structure of DHiF}
To extract finer-grained characteristics, we design DHiF to be locally adaptive as shown in Fig. \ref{DHiF}. Each local region corresponds to a dedicated filter bank to learn discriminative representations. The input of DHiF is the features $\boldsymbol{F}_{in}\in \mathbb{R}^{C\times H\times W}$ of an infrared image. Its output is the features $\boldsymbol{F}_{out}\in \mathbb{R}^{C^{\prime}\times H^{\prime}\times W^{\prime}}$ further extracted from the intermediate filtered features generated during the forward pass. The operation details of DHiF are described below.

As shown in Fig. \ref{DHiF}, the input features are first sent to a normalization layer for the single-channel feature map $\boldsymbol{F}_{norm}\in \mathbb{R}^{H\times W}$. Then, a series of operations is applied to the normalized feature map to generate dynamic local filters. Initially, the local normalized feature $\boldsymbol{F}_{norm}^{local}\in \mathbb{R}^{k\times k}$ within the sliding window is flattened into a 1-D vector, i.e., $\mathbb{R}^{k\times k} \rightarrow \mathbb{R}^{1\times k^2}$. Subsequently, a linear projection $\rho(\cdot)$ is applied to this normalized vector, yielding $k^2$ vectors (i.e., $\boldsymbol{v}_1,\boldsymbol{v}_2, ..., \boldsymbol{v}_{k^2} \in \mathbb{R}^{k^2\times 1}$) that encode the grayscale variation characteristics of the local region. Through a non-linear mapping $\theta(\cdot)$, all elements of the vectors are projected onto interval $[-1, 1]$, which makes DHiF sensitive to HFCs. Then, $k^2$ region-specific filters $\boldsymbol{f}=[f_1, f_2, ..., f_{k^2}]$ are generated. The process can be described as
\begin{equation}
    \label{equ_filters1}
    \boldsymbol{v}_1,\boldsymbol{v}_2, ..., \boldsymbol{v}_{k^2} = \rho\big(vec(\boldsymbol{F}_{norm}^{local})\big),
\end{equation}
\begin{equation}
    \label{equ_filters2}
    \boldsymbol{W}_{f_1},\boldsymbol{W}_{f_2}, ..., \boldsymbol{W}_{f_{k^2}} = \theta(\boldsymbol{v}_1,\boldsymbol{v}_2, ..., \boldsymbol{v}_{k^2}),
\end{equation}
where $vec(\cdot)$ and $\theta(\cdot)$ denote the flattening operation and the nonlinear hyperbolic tangent mapping function (i.e., $tanh(\cdot)$), respectively. $\boldsymbol{W}_{f_i}\in \mathbb{R}^{k^2\times 1}$ represents the kernel of the $i^{th}$ filter. $\boldsymbol{W}_{\boldsymbol{f}} \in \mathbb{R}^{k^2\times k^2}$ is all kernels of $\boldsymbol{f}$.

To obtain the filtered features $\boldsymbol{F}_{filt}^{local}\in \mathbb{R}^{C\times k^2}$ of the local region, flatten the original local input features $\boldsymbol{F}_{in}^{local}\in \mathbb{R}^{C\times k\times k}\rightarrow \mathbb{R}^{C\times k^2}$, and then, apply all filters of this region to each channel of the flattened features, i.e.,
\begin{equation}
    \label{equ_filters2}
    \boldsymbol{F}_{filt}^{local} =vec(\boldsymbol{F}_{in}^{local})\boldsymbol{W}_{\boldsymbol{f}}.
\end{equation}
The filtered features record distinctive information of HFCs in different local regions.

Afterwards, $C^{\prime}$ standard convolutions with the weight matrix $\boldsymbol{W}_j\in \mathbb{R}^{C\times k^2}$ are performed on the combination of local input features and its filtered features to obtain the output features $\boldsymbol{F}_{out}^{local} \in \mathbb{R}^{C^{\prime}\times 1\times 1}$ of the local region. This process can be described as
\begin{equation}
    \label{equ_filters2}
    \boldsymbol{F}_{out}^{local}[j] = \sum\limits_{1}^{k^2} \Big( \big( vec(\boldsymbol{F}_{in}^{local}) + \boldsymbol{F}_{filt}^{local}\big) \odot \boldsymbol{W}_j \Big),
\end{equation}
where $\odot$ denotes the Hadamard product.

Finally, after processing all local regions in the input feature map, the output feature map $\boldsymbol{F}_{out}$ of DHiF is finally obtained. In this map, the HFCs are selectively focused, and their discriminative representations are preliminarily learned.

DHiF can function as a drop-in replacement for standard convolution in SIRST detection networks, and be embedded into the residual block \cite{he2016deep} for a new dynamic high-frequency residual block (DHiF-Res block, details are given in the supplemental material).

\section{Experiments}

\subsection{Experimental Setup}

\subsubsection{Datasets}
We evaluated our method on real-scene datasets (the IRSDT-1k \cite{zhang2022isnet} and NUAA-SIRST \cite{dai2021asymmetric} datasets).

\subsubsection{Metrics}
We followed most segmentation-based SIRST detection networks \cite{yuan2024sctransnet, li2025ilnet} to use intersection over union ($IoU$), normalized $IoU$ ($nIoU$), probability of detection ($P_d$), and false alarm rate ($F_a$) as quantitative metrics for performance evaluation, and FPS and the number of parameters (\#Params) for computational efficiency evaluation. Detailed descriptions of these metrics are provided in the supplementary material. In the following results, $IoU$, $nIoU$, and $P_d$ are reported in units of $10^{-2}$, while $F_a$ uses units of $10^{-5}$.

\begin{table}[t!]
\caption{Detection performance of different convolutions in various networks' encoders. Best results are in \textbf{bold}, and second-best results are \underline{underlined}, versus standard convolution results.}
\label{tab:SOTA2}
\centering
\begin{threeparttable}
\renewcommand{\arraystretch}{0.96}
\setlength{\tabcolsep}{1.6pt}{
\begin{tabular}{|c|cc cccc cccc|}
\hline
\multicolumn{2}{|c}{\multirow{2}{*}{Networks}} & \multirow{2}{*}{Convs} & \multicolumn{4}{c}{IRSTD-1K} & \multicolumn{4}{c|}{NUAA-SIRST} \\
\cline{4-11}
\multicolumn{2}{|c}{} &  & $IoU$ & $nIoU$ & $P_d$ & $F_a$ & $IoU$ & $nIoU$ & $P_d$ & $F_a$ \\
\hline
\multirow{20}*{\rotatebox{90}{CNN}}
 & \multirow{5}{*}{FC3Net} & Standard & 60.12 & 59.36 & 89.23 & 4.73 & 31.09 & 30.30 & \textbf{94.27} & 30.19 \\
 & & CDC & 57.99 & 56.06 & \textbf{92.93} & 5.20 & 32.38 & 31.26 & 92.37 & 26.40 \\
 & & WTConv & 61.55 & 60.67 & 89.90 & 3.10 & 70.17 & 71.25 & 92.75 & \textbf{3.48} \\
 & & PConv & \textbf{62.75} & \underline{61.02} & 90.24 & \textbf{2.37} & \underline{70.81} & \underline{72.12} & \textbf{94.27} & 5.63 \\
 & & DHiF & \underline{62.10} & \textbf{61.08} & \underline{91.25} & \underline{2.49} & \textbf{71.33} & \textbf{72.54} & \textbf{94.27} & \underline{3.77} \\
\cline{2-11}
 & \multirow{5}{*}{DNANet} & Standard & 63.27 & 63.56 & 88.55 & \underline{1.25} & 71.97 & \underline{78.66} & 95.04 & 6.31 \\
 & & CDC & 65.27 & 65.01 & \textbf{90.91} & 1.62 & 75.41 & 78.49 & \textbf{96.18} & 2.96 \\
 & & WTConv & 65.17 & \underline{65.11} & 90.24 & \textbf{0.90} & 74.82 & 77.45 & 93.13 & 1.93 \\
 & & PConv & \underline{66.03} & 64.83 & 90.24 & 1.83 & \underline{75.64} & 78.08 & 92.75 & \underline{1.63} \\
 & & DHiF & \textbf{69.42} & \textbf{65.68} & \underline{90.57} & 1.38 & \textbf{77.68} & \textbf{79.03} & \underline{95.42} & \textbf{1.53} \\
\cline{2-11}
 & \multirow{5}{*}{ISNet} & Standard & 62.38 & 62.25 & 89.90 & 3.27 & \underline{71.10} & \underline{72.88} & \textbf{95.04} & 5.52 \\
 & & CDC & \underline{63.65} & \textbf{64.27} & \underline{92.93} & \textbf{2.51} & 70.18 & 70.68 & \underline{93.13} & \underline{5.23} \\
 & & WTConv & 58.68 & 57.25 & \textbf{93.27} & 4.50 & 36.47 & 47.28 & 88.93 & 51.00 \\
 & & PConv & 54.43 & 55.51 & 91.25 & 6.80 & 45.92 & 45.68 & 88.93 & 15.76 \\
 & & DHiF & \textbf{68.08} & \underline{63.90} & 90.91 & \underline{2.92} & \textbf{72.57} & \textbf{75.27} & \underline{93.13} & \textbf{3.89} \\
\cline{2-11}
 & \multirow{5}{*}{MSHNet} & Standard & 63.56 & 64.79 & \underline{89.90} & 1.22 & \underline{76.98} & \textbf{77.04} & \underline{95.80} & 1.97 \\
 & & CDC & \underline{64.53} & \underline{65.28} & 89.23 & \underline{0.97} & 73.35 & 76.26 & 93.13 & \underline{1.30} \\
 & & WTConv & 64.20 & 64.15 & 87.88 & \textbf{0.56} & 74.47 & 73.56 & 92.37 & 1.56 \\
 & & PConv & 63.16 & 65.02 & \underline{89.90} & 2.36 & 74.86 & 74.25 & 94.27 & 1.56 \\
 & & DHiF & \textbf{65.15} & \textbf{65.50} & \textbf{90.91} & 1.10 & \textbf{77.07} & 76.13 & \textbf{96.56} & \textbf{1.08} \\
\hline
\multirow{15}*{\rotatebox{90}{CNN-T}}
 & \multirow{5}{*}{MTU-Net} & Standard & \underline{64.78} & \underline{64.53} & 88.22 & \textbf{1.51} & \underline{75.73} & \underline{78.71} & \textbf{94.66} & 3.09 \\
 & & CDC & 64.07 & 64.12 & 91.25 & 1.80 & 73.57 & 75.13 & 93.89 & 3.98 \\
 & & WTConv & 63.80 & 63.60 & \underline{91.92} & 1.94 & 74.44 & 76.76 & 94.27 & 3.25 \\
 & & PConv & 63.70 & 64.20 & \textbf{92.26} & 2.89 & 75.15 & 76.96 & 94.27 & \textbf{2.09} \\
 & & DHiF & \textbf{65.56} & \textbf{65.50} & 89.23 & 1.54 & \textbf{77.12} & \textbf{79.19} & \textbf{94.66} & \underline{2.48} \\
\cline{2-11}
 & \multirow{5}{*}{SCTransNet} & Standard & 64.05 & 65.10 & 90.24 & \underline{2.27} & \underline{77.97} & \underline{79.39} & 95.04 & \underline{2.86} \\
 & & CDC & \underline{66.31} & 65.39 & \underline{92.26} & 2.60 & 73.94 & 77.00 & \underline{95.42} & 4.49 \\
 & & WTConv & \textbf{66.52} & \textbf{65.74} & 90.57 & 2.30 & 72.63 & 76.32 & \textbf{96.56} & 4.92 \\
 & & PConv\tnote{1} & 0.04 & 0.04 & - & - & 0.01 & 0.01 & - & - \\
 & & DHiF & 65.58 & \underline{65.47} & \textbf{92.59} & \textbf{2.17} & \textbf{78.83} & \textbf{79.40} & \textbf{96.56} & \textbf{2.60} \\
\cline{2-11}
 & \multirow{5}{*}{APTNet} & Standard & 63.18 & 63.89 & 88.22 & \textbf{1.02} & \underline{75.08} & \underline{77.57} & \textbf{96.95} & 3.22 \\
 & & CDC & \textbf{66.38} & \textbf{66.08} & \textbf{90.57} & 1.99 & 71.09 & 75.09 & 92.37 & \underline{2.42} \\
 & & WTConv & 64.26 & 62.82 & 86.20 & 1.03 & 71.42 & 75.08 & 91.22 & 2.78 \\
 & & PConv & 61.28 & 61.66 & 88.55 & 1.28 & 71.05 & 76.05 & 93.89 & 6.15 \\
 & & DHiF & \underline{65.25} & \underline{64.17} & \underline{89.23} & 1.92 & \textbf{77.79} & \textbf{78.29} & \textbf{96.95} & \textbf{1.63} \\
\hline
\end{tabular}}
\begin{tablenotes}
\footnotesize
\item[1] SCTransNet with PConv fails to converge during the training process.
\end{tablenotes}
\end{threeparttable}
\end{table}

\begin{figure*}[t]
\centering
\includegraphics[width=1\textwidth]{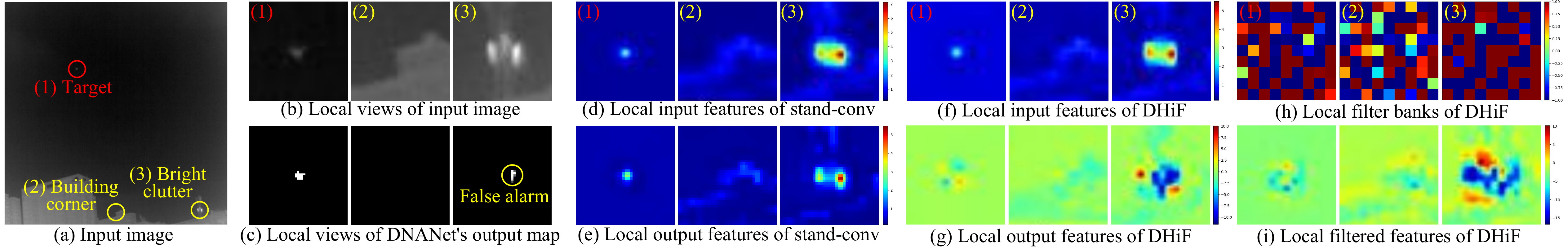}
\caption{Visualizations of features in the third levels of DNANet variants without and with DHiF. (a) The input image including (1) target, (2) building corner, and (3) bright clutter. (c) Local view of the output map generated by the original DNANet. (d, f) Local input features and (e, g) local output features of the first convolutions (standard convolution and DHiF). (h) Local filter banks of three HFCs.}
\label{vis_conv}
\end{figure*}

\subsection{Comparison to the State-of-the-arts}
To demonstrate the generality and superiority of DHiF for SIRST detection, we compared DHiF with different convolutions, including CDC \cite{zhang2024mdigcnet} for high frequency focus, WTConv \cite{finder2024wavelet} for large receptive fields, and PConv \cite{yang2025pinwheel} for spatial characteristics of the SIRST pixel distribution. Specifically, we substitute different convolutions for standard convolutions in encoders of different SIRST detection networks, including CNN-based methods FC3Net \cite{zhang2022exploring}, DNANet \cite{li2022dense}, ISNet \cite{zhang2022isnet} and MSHNet \cite{liu2024infraredMSHNet}, and CNN-Transformer hybrid (CNN-T) methods MTU-Net \cite{wu2023mtu}, SCTransNet \cite{yuan2024sctransnet}, and APTNet \cite{zhang2025aptnet}. The quantitative results are presented in Table \ref{tab:SOTA2}, and the qualitative results are shown in the supplementary material.

Compared with the original networks, the DHiF-enhanced variants show obvious quantitative and qualitative improvements in detection performance on both datasets. Specifically, the enhanced models produce more accurate segmentations and better distinguish targets from complex backgrounds. This stems from DHiF's capability to finely and differentially capture high-frequency information, which enables superior detection efficacy universally.

Compared with other specially designed convolutions, DHiF demonstrates superior generalizability, consistently improving performance across different networks and datasets. On the IRSTD‑1k dataset, CDC, WTConv and PConv degrade detection performance obviously in certain networks (e.g., FC3Net, ISNet and SCTransNet), indicating their limited adaptability to diverse networks. On the NUAA-SIRST dataset, all state-of-the-art convolutions (except DHiF) induce severe performance degradation in most networks. In contrast, DHiF consistently achieves robust performance across different datasets and facilitates obvious performance gains to various networks. This highlights the superior adaptability of DHiF to various scenarios and targets. The adaptability comes from the capability of DHiF to dynamically generate specialized filter operators based on local regions in various scenes. Dynamic local filter operators can focus differentially on high-frequency regions in infrared images, thereby enhancing the robustness of SIRST detection networks.

\subsection{Ablation Study}

\subsubsection{Position of drop-in replacement and efficiency}
To investigate whether arbitrary convolutions in encoders are suitable for replacement, we replaced residual blocks at different levels of the DNANet \cite{li2022dense} encoder with DHiF-Res blocks (level 1 is the input layer; level 5 is the encoder output layer). We compared the detection performance and computational efficiency across these variants as shown in Table \ref{tab:level}. Additional ablation study on DHiF is in the supplementary material.

\begin{table}[t]
    \caption{Detection results achieved by variants of DNANet with DHiFs in different levels on the NUAA-SIRST dataset.}
    \label{tab:level}
    \centering
    \renewcommand{\arraystretch}{0.9}
    \setlength{\tabcolsep}{5pt}{
    \begin{tabular}{|c c c c c c c|}
    \hline
    level & $IoU$ & $nIoU$ & $P_d$ & $F_a$ & FPS & \#Params \\
    \hline
    None & 71.97 & 78.66 & 95.04 & 6.31 & 8.53 & 18346 K \\
    $1^{st}$ (1 block) & 72.95 & 77.77 & 94.66 & 4.89 & 8.43 & 18349 K \\
    $2^{nd}$ (2 blocks) & 78.37 & 79.29 & 96.18 & 3.32 & 8.05 & 18352 K \\
    $3^{rd}$ (2 blocks) & 77.02 & 78.99 & 95.42 & 2.57 & 8.42 & 18352 K \\
    $4^{th}$ (2 blocks) & 77.19 & 78.01 & 96.18 & 2.68 & 8.47 & 18352 K \\
    $5^{th}$ (2 blocks) & 75.48 & 79.10 & 96.18 & 3.86 & 8.51 & 18352 K \\
    \hline
    \end{tabular}}
\end{table}

The results show that using DHiF in the $2^{nd}$ to $5^{th}$ levels yields an obvious improvement in detection performance ($IoU$ increases by $5.05\%$, $nIoU$ by $0.19\%$, $P_d$ by $0.95\%$, and $F_a$ decreases by $3.20\times 10^{-5}$ on average), while having only a negligible impact on computational efficiency. In contrast, applying DHiF at the input layer provides no gain and even reduces $P_d$ ($\downarrow0.38\%$). This indicates that adaptive local filters are unsuitable as an input layer operating directly on raw infrared images, but work effectively in hidden layers processing feature maps. We suggest that it is because, as the input layer, such filters may disrupt the original image distribution and thereby interfere with subsequent feature extraction. Therefore, replacing standard convolutions with DHiF in encoder hidden layers can enhance SIRST detection performance without significantly affecting computational efficiency.

\subsubsection{Effectiveness on learning differential representations}
To illustrate how DHiF facilitates discriminative representation learning, we visualized the differences between the feature maps with and without DHiF, as well as the local filter banks generated for distinct HFCs. Specifically, we replaced the residual blocks in the second and third levels of DNANet with DHiF-Res blocks, and the visualizations are presented in Fig. \ref{vis_conv} and in the supplementary material, respectively.

As shown in Fig. \ref{vis_conv}(e), DNANet without DHiF produces consistent positive responses in the target, building edge, and bright clutter areas, causing the false alarm. By contrast, DNANet with DHiF responds differentially to distinct HFCs (Fig. \ref{vis_conv}(g)). In bright clutter areas, most pixels yield strong negative responses, while a narrow edge-adjacent zone retains positive activations. This contrasting behavior within local regions highlights DHiF's ability to capture fine-grained details. It further demonstrates that DHiF excels at discriminatively representing different HFCs, thereby enabling networks to better distinguish targets from other HFCs.

\section{Conclusion}
In this paper, we emphasize the importance of discriminative representation learning of different HFCs and propose a novel dynamic high-frequency convolution (DHiF) to enhance this learning. DHiF is designed as a drop-in replacement for standard convolution, which can be equipped with arbitrary SIRST detection networks with convolutional structures to improve detection performance. It consists of a region-specific dynamic filter generator and a standard convolution to capture the distinctive grayscale variation characteristics and learn the discriminative information of different HFC regions. Extensive experiments on various state-of-the-art networks and real datasets are conducted to verify the effectiveness and generality of our DHiF.


\begin{thebibliography}{10}
\providecommand{\url}[1]{#1}
\csname url@samestyle\endcsname
\providecommand{\newblock}{\relax}
\providecommand{\bibinfo}[2]{#2}
\providecommand{\BIBentrySTDinterwordspacing}{\spaceskip=0pt\relax}
\providecommand{\BIBentryALTinterwordstretchfactor}{4}
\providecommand{\BIBentryALTinterwordspacing}{\spaceskip=\fontdimen2\font plus
\BIBentryALTinterwordstretchfactor\fontdimen3\font minus
  \fontdimen4\font\relax}
\providecommand{\BIBforeignlanguage}[2]{{%
\expandafter\ifx\csname l@#1\endcsname\relax
\typeout{** WARNING: IEEEtran.bst: No hyphenation pattern has been}%
\typeout{** loaded for the language `#1'. Using the pattern for}%
\typeout{** the default language instead.}%
\else
\language=\csname l@#1\endcsname
\fi
#2}}
\providecommand{\BIBdecl}{\relax}
\BIBdecl
\bibitem{zhang2022exploring}
M.~Zhang, K.~Yue, J.~Zhang, Y.~Li, and X.~Gao, ``Exploring feature compensation
  and cross-level correlation for infrared small target detection,'' in
  \emph{ACM MM}, 2022.

\bibitem{zhang2023dim2clear}
M.~Zhang, R.~Zhang, J.~Zhang, J.~Guo, Y.~Li, and X.~Gao, ``{Dim2Clear} network
  for infrared small target detection,'' \emph{IEEE TGRS}, 2023.

\bibitem{ribeiro2017data}
R.~Ribeiro, G.~Cruz, J.~Matos, and A.~Bernardino, ``A data set for airborne
  maritime surveillance environments,'' \emph{IEEE TCSVT}, 2017.

\bibitem{zhou2019background}
A.~Zhou, W.~Xie, and J.~Pei, ``Background modeling in the fourier domain for
  maritime infrared target detection,'' \emph{IEEE TCSVT}, 2019.

\bibitem{huang2023anti}
B.~Huang, J.~Li, J.~Chen, G.~Wang, J.~Zhao, and T.~Xu, ``{Anti-UAV410}: A
  thermal infrared benchmark and customized scheme for tracking drones in the
  wild,'' \emph{IEEE TPAMI}, 2023.

\bibitem{li2025probing}
R.~Li, W.~An, X.~Ying, Y.~Wang, Y.~Dai, L.~Wang, M.~Li, Y.~Guo, and L.~Liu,
  ``Probing deep into temporal profile makes the infrared small target detector
  much better,'' \emph{arXiv preprint arXiv:2506.12766}, 2025.

\bibitem{zhang2022rkformer}
M.~Zhang, H.~Bai, J.~Zhang, R.~Zhang, C.~Wang, J.~Guo, and X.~Gao,
  ``{RKformer}: {Runge-Kutta} transformer with random-connection attention for
  infrared small target detection,'' in \emph{ACM MM}, 2022.

\bibitem{zhang2024irprunedet}
M.~Zhang, H.~Yang, J.~Guo, Y.~Li, X.~Gao, and J.~Zhang, ``{IRPruneDet}:
  Efficient infrared small target detection via wavelet structure-regularized
  soft channel pruning,'' in \emph{AAAI}, 2024.

\bibitem{li2023direction}
R.~Li, W.~An, C.~Xiao, B.~Li, Y.~Wang, M.~Li, and Y.~Guo, ``Direction-coded
  temporal {U-shape} module for multiframe infrared small target detection,''
  \emph{IEEE TNNLS}, 2023.

\bibitem{guo2024dmfnet}
T.~Guo, B.~Zhou, F.~Luo, L.~Zhang, and X.~Gao, ``{DMFNet}: Dual-encoder
  multistage feature fusion network for infrared small target detection,''
  \emph{IEEE TGRS}, 2024.

\bibitem{li2024multi}
G.~Li, J.~Zhang, E.~Yang, H.~Jiang, and D.~Zeng, ``Multi-level information
  fusion network with edge information injection for single-band cloud
  detection,'' \emph{IEEE TCSVT}, 2024.

\bibitem{zeng2024mmi}
Y.~Zeng, T.~Liang, Y.~Jin, and Y.~Li, ``{MMI-Det}: Exploring multi-modal
  integration for visible and infrared object detection,'' \emph{IEEE TCSVT},
  2024.

\bibitem{wei2016multiscale}
Y.~Wei, X.~You, and H.~Li, ``Multiscale patch-based contrast measure for small
  infrared target detection,'' \emph{PR}, 2016.

\bibitem{zhu2020balanced}
H.~Zhu, J.~Zhang, G.~Xu, and L.~Deng, ``Balanced ring top-hat transformation
  for infrared small-target detection with guided filter kernel,'' \emph{IEEE
  TAES}, 2020.

\bibitem{wu2023mtu}
T.~Wu, B.~Li, Y.~Luo, Y.~Wang, C.~Xiao, T.~Liu, J.~Yang, W.~An, and Y.~Guo,
  ``{MTU-Net}: Multi-level {TransUNet} for space-based infrared tiny ship
  detection,'' \emph{IEEE TGRS}, 2023.

\bibitem{guo2025infrared}
T.~Guo, B.~Zhou, F.~Luo, and L.~Zhang, ``Infrared small target detection via
  diverse features harmonization,'' \emph{IEEE TGRS}, 2025.

\bibitem{zhao2025chirplet}
X.~Zhao, Q.~Ming, Y.~Yang, W.~Hu, W.~Li, and R.~Tao, ``Chirplet fourier
  analysis network for cross-scene classification of multisource remote sensing
  data,'' \emph{IEEE TGRS}, 2025.

\bibitem{wang2025hypersigma}
D.~Wang, M.~Hu, Y.~Jin, Y.~Miao, J.~Yang, Y.~Xu, X.~Qin, J.~Ma, L.~Sun, C.~Li
  \emph{et~al.}, ``{HyperSIGMA}: Hyperspectral intelligence comprehension
  foundation model,'' \emph{IEEE TPAMI}, 2025.

\bibitem{liu2024spatial}
S.~Liu, C.~Fu, Y.~Duan, X.~Wang, and F.~Luo, ``Spatial-spectral enhancement and
  fusion network for hyperspectral image classification with few labeled
  samples,'' \emph{IEEE TGRS}, 2024.

\bibitem{liuRandomFeatures}
L.~Liu and P.~Fieguth, ``Texture classification from random features,''
  \emph{IEEE TPAMI}, 2012.

\bibitem{chen2025event}
N.~Chen, C.~Zhang, W.~An, L.~Wang, M.~Li, and Q.~Ling, ``Event-based motion
  deblurring with blur-aware reconstruction filter,'' \emph{IEEE TCSVT}, 2025.

\bibitem{yang2023small}
P.~Yang, L.~Dong, and W.~Xu, ``Small maritime target detection using gradient
  vector field characterization of infrared image,'' \emph{IEEE JSTARS}, 2023.

\bibitem{gu2021interpreting}
J.~Gu and C.~Dong, ``Interpreting super-resolution networks with local
  attribution maps,'' in \emph{CVPR}, 2021.

\bibitem{li2022dense}
B.~Li, C.~Xiao, L.~Wang, Y.~Wang, Z.~Lin, M.~Li, W.~An, and Y.~Guo, ``Dense
  nested attention network for infrared small target detection,'' \emph{IEEE
  TIP}, 2022.

\bibitem{li2025ilnet}
H.~Li, J.~Yang, R.~Wang, and Y.~Xu, ``{ILNet}: Low-level matters for salient
  infrared small target detection,'' \emph{IEEE TAES}, 2025.

\bibitem{yuan2024sctransnet}
S.~Yuan, H.~Qin, X.~Yan, N.~Akhtar, and A.~Mian, ``{SCTransNet}:
  Spatial-channel cross transformer network for infrared small target
  detection,'' \emph{IEEE TGRS}, 2024.

\bibitem{he2016deep}
K.~He, X.~Zhang, S.~Ren, and J.~Sun, ``Deep residual learning for image
  recognition,'' in \emph{CVPR}, 2016.

\bibitem{zhang2022isnet}
M.~Zhang, R.~Zhang, Y.~Yang, H.~Bai, J.~Zhang, and J.~Guo, ``{ISNet}: Shape
  matters for infrared small target detection,'' in \emph{CVPR}, 2022.

\bibitem{dai2021asymmetric}
Y.~Dai, Y.~Wu, F.~Zhou, and K.~Barnard, ``Asymmetric contextual modulation for
  infrared small target detection,'' in \emph{WACV}, 2021.

\bibitem{zhang2024mdigcnet}
L.~Zhang, J.~Luo, Y.~Huang, F.~Wu, X.~Cui, and Z.~Peng, ``{MDIGCNet}:
  Multi-directional information-guided contextual network for infrared small
  target detection,'' \emph{IEEE JSTARS}, 2024.

\bibitem{finder2024wavelet}
S.~E. Finder, R.~Amoyal, E.~Treister, and O.~Freifeld, ``Wavelet convolutions
  for large receptive fields,'' in \emph{ECCV}, 2024.

\bibitem{yang2025pinwheel}
J.~Yang, S.~Liu, J.~Wu, X.~Su, N.~Hai, and X.~Huang, ``Pinwheel-shaped
  convolution and scale-based dynamic loss for infrared small target
  detection,'' in \emph{AAAI}, 2025.

\bibitem{liu2024infraredMSHNet}
Q.~Liu, R.~Liu, B.~Zheng, H.~Wang, and Y.~Fu, ``Infrared small target detection
  with scale and location sensitivity,'' in \emph{CVPR}, 2024.

\bibitem{zhang2025aptnet}
Y.~Zhang, W.~Bao, W.~Wan, Q.~Xiao, Y.~Tang, X.~Zou, L.~Huang, K.~Zhong, and
  Y.~Lan, ``{APTNet}: Adaptive partial transformer network for infrared small
  target detection,'' \emph{IEEE Sens. J.}, 2025.

\end{thebibliography}

\begin{IEEEbiography}[{\includegraphics[width=1in,height=1.25in,clip,keepaspectratio]{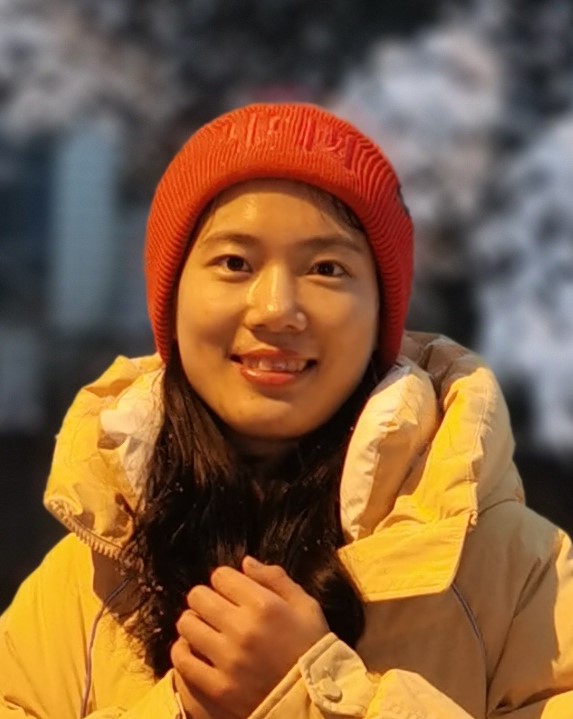}}]{Ruojing Li}
received the B.E. degree in electronic engineering from the National University of Defense Technology (NUDT), Changsha, China, in 2020. She is currently working toward the Ph.D. degree in information and communication engineering from NUDT. Her research interests include infrared small target detection, particularly on multi-frame detection, deep learning. For more information, please visit \href{https://tinalrj.github.io/}{https://tinalrj.github.io/}.
\end{IEEEbiography}

\begin{IEEEbiography}[{\includegraphics[width=1in,height=1.25in,clip,keepaspectratio]{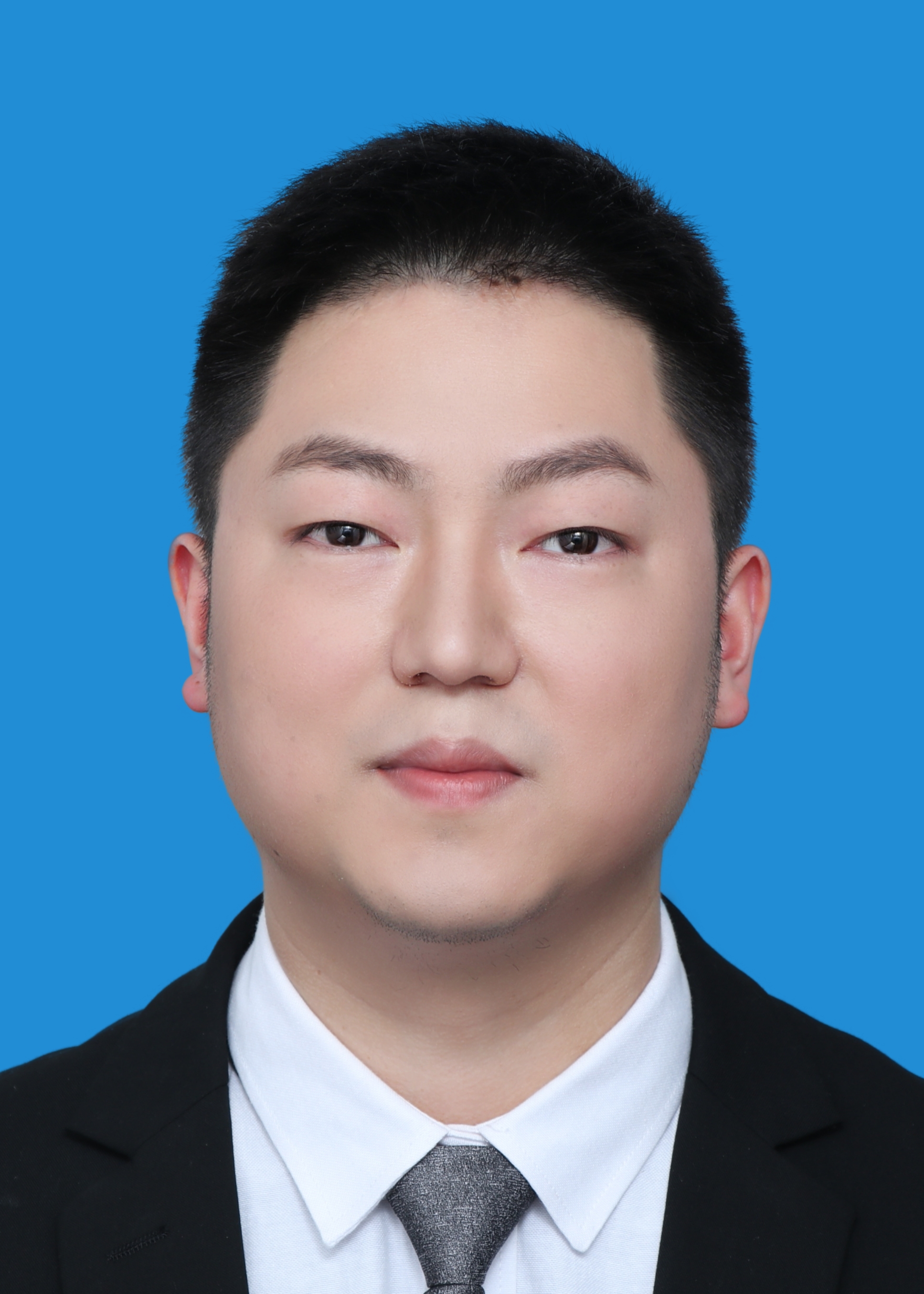}}]{Chao Xiao}
received the B.E., M.E., and Ph.D. degrees from the National University of Defense Technology (NUDT), Changsha, China, in 2016, 2018, and 2023, respectively. He is currently a Lecturer with the College of Electronic Science and Technology, NUDT. His recent research interests focus on small target detection and tracking in satellite videos.
\end{IEEEbiography}

\begin{IEEEbiography}[{\includegraphics[width=1in,height=1.25in,clip,keepaspectratio]{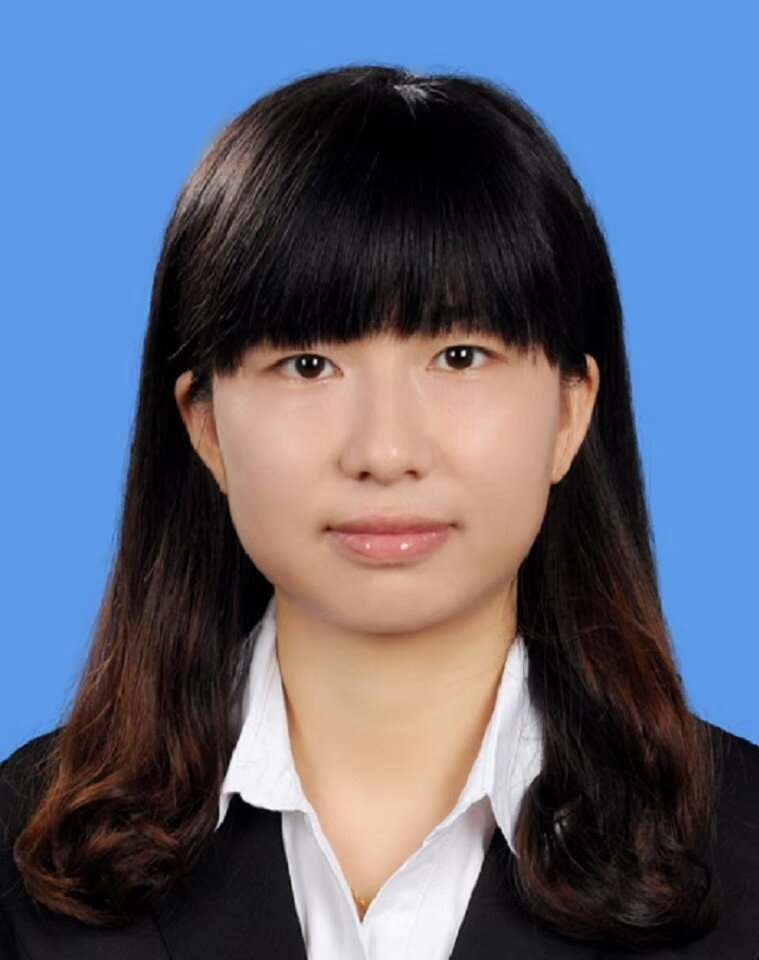}}]{Qian Yin}
received the B.E. degree in Electrical Engineering from Hunan University of Science and Technology (HNUST), Xiangtan, China, in 2014, and the M.E. degree in Optical Engineering from South China Normal University (SCNU), Guangzhou, China, in 2016. She is currently pursuing the Ph.D. degree with the College of Electronic Science and Technology, National University of Defense Technology (NUDT). Her research interests focus on object detection and tracking.
\end{IEEEbiography}

\begin{IEEEbiography}[{\includegraphics[width=1in,height=1.25in,clip,keepaspectratio]{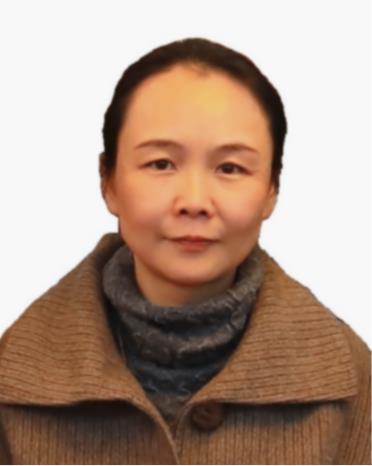}}]{Wei An}
received the Ph.D. degree from the National University of Defense Technology (NUDT), Changsha, China, in 1999. She was a Senior Visiting Scholar with the University of Southampton, Southampton, U.K., in 2016. She is currently a Professor with the College of Electronic Science and Technology, NUDT. She has authored or co-authored over 100 journal and conference publications. Her current research interests include signal processing and image processing.
\end{IEEEbiography}

\begin{IEEEbiography}[{\includegraphics[width=1in,height=1.25in,clip,keepaspectratio]{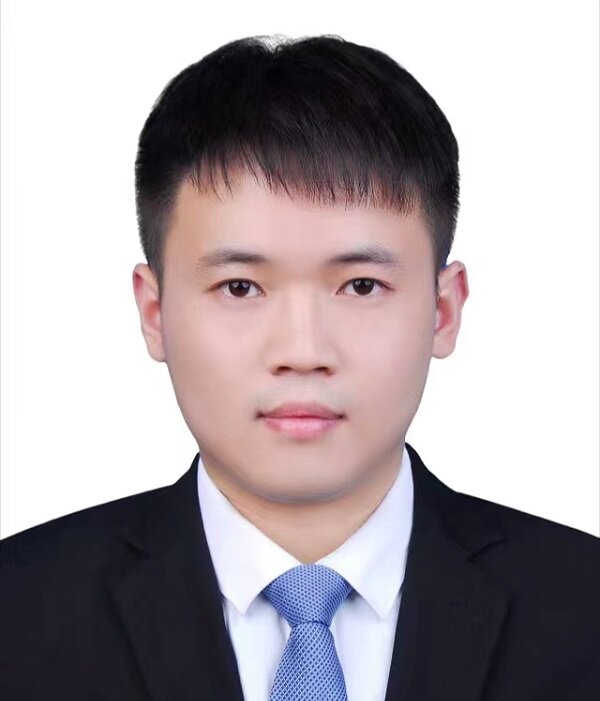}}]{Nuo Chen}
received the B.E. degree in Mechanical Design manufacture and Automation and M.S. degree in Mechanical and Electrical Engineering from the Central South University (CSU), China, in 2019 and 2022, respectively. He is currently working toward the Ph.D. degree in Information and Communication Engineering from National University of Defense Technology (NUDT), Changsha, China. His research interests include image processing, neuromorphic cameras and computer vision.
\end{IEEEbiography}

\begin{IEEEbiography}[{\includegraphics[width=1in,height=1.25in,clip,keepaspectratio]{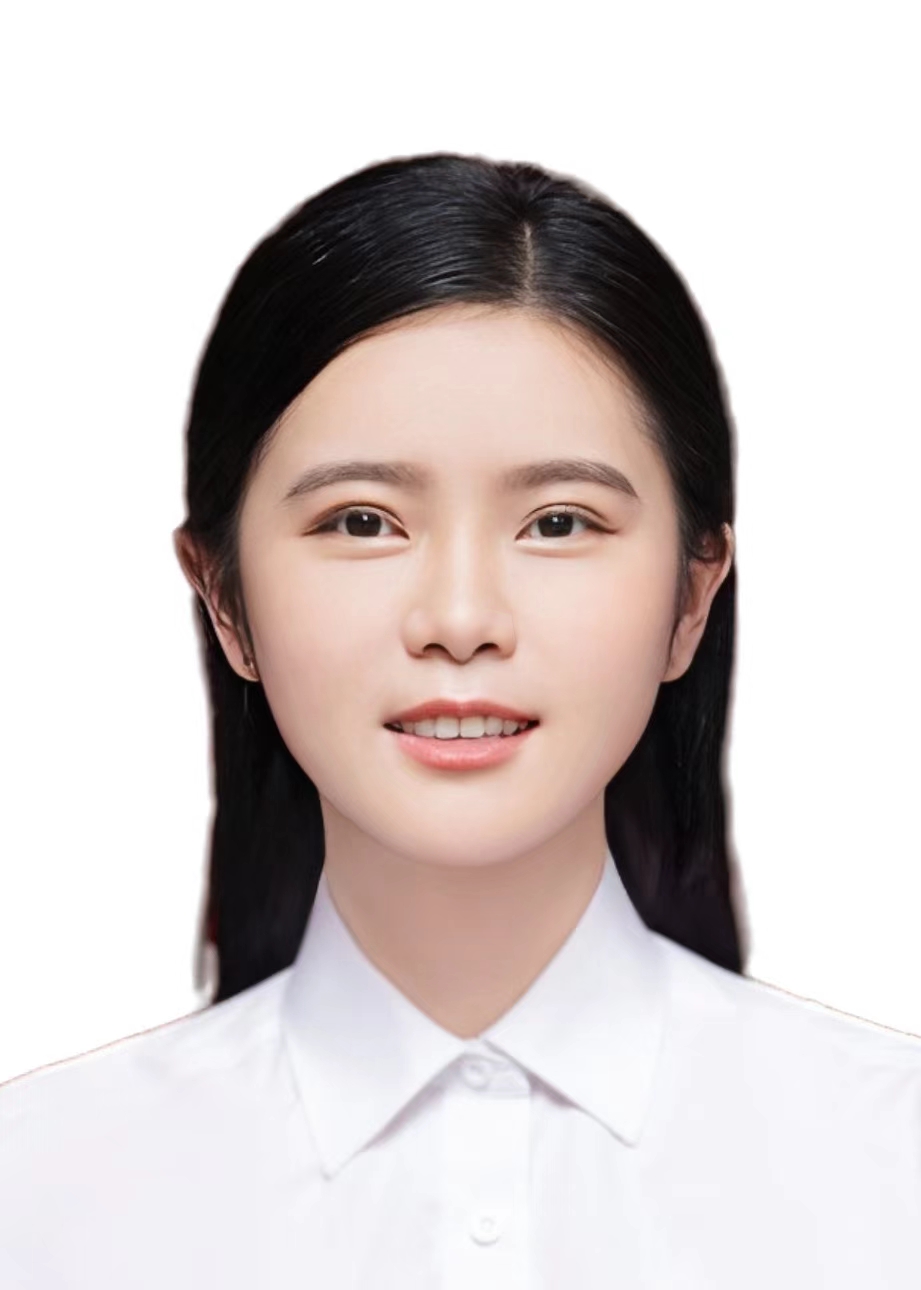}}]{Xinyi Ying}
received the Ph.D. degree in information and communication engineering from the National University of Defense Technology (NUDT), Changsha, China, in 2024. She is currently a Lecturer with the College of Electronic Science and Technology, NUDT. Her research interests focus on space-based optical surveillance, and detection and tracking of infrared small targets.
\end{IEEEbiography}

\begin{IEEEbiography}[{\includegraphics[width=1in,height=1.25in,clip,keepaspectratio]{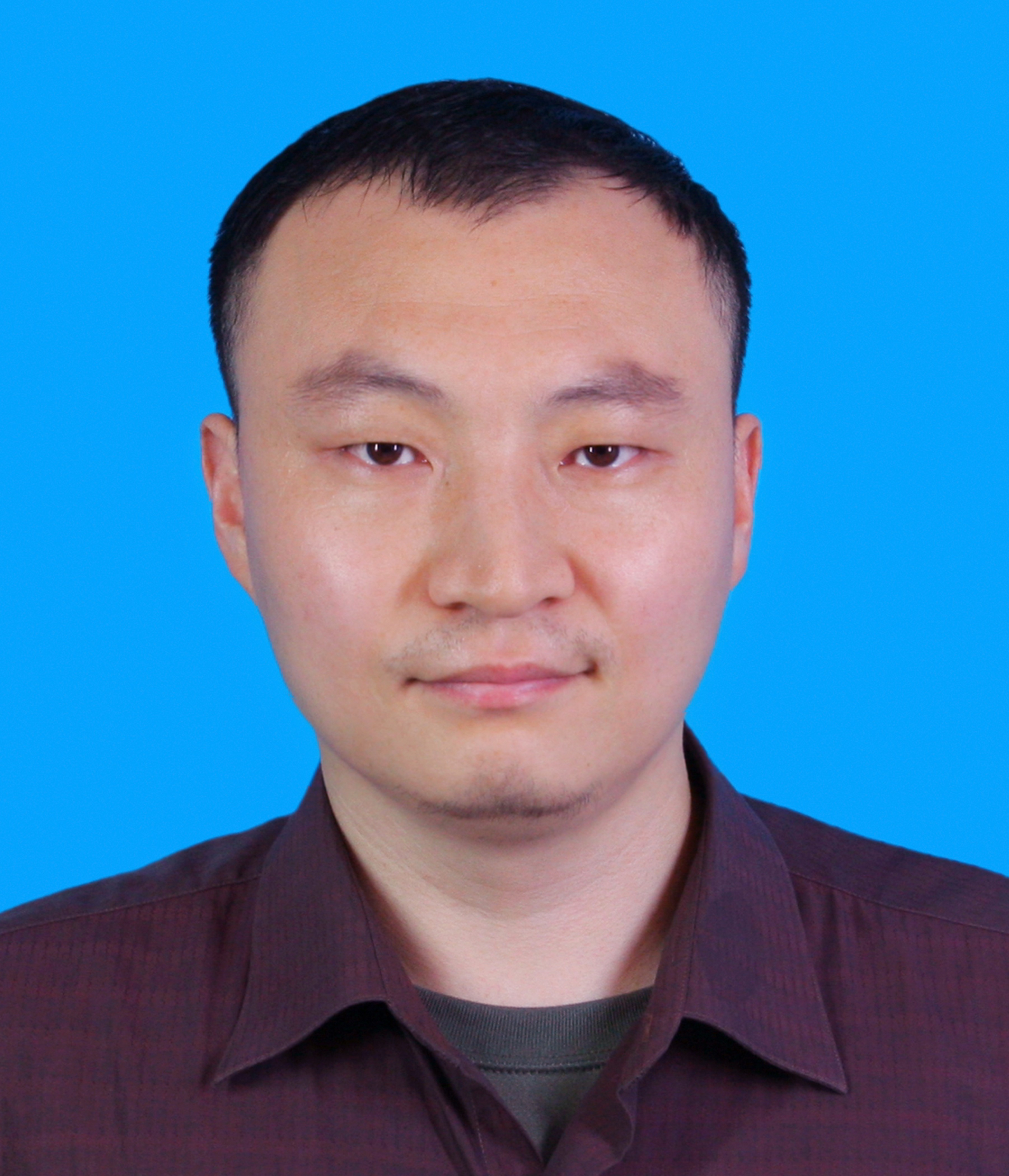}}]{Miao Li}
received the M.E. and Ph.D. degrees from the National University of Defense Technology (NUDT) in 2012 and 2017, respectively. He is currently an Associate Professor with the College of Electronic Science and Technology, NUDT. His current research interests include infrared dim and small target detection and event detection.
\end{IEEEbiography}

\begin{IEEEbiography}[{\includegraphics[width=1in,height=1.25in,clip,keepaspectratio]{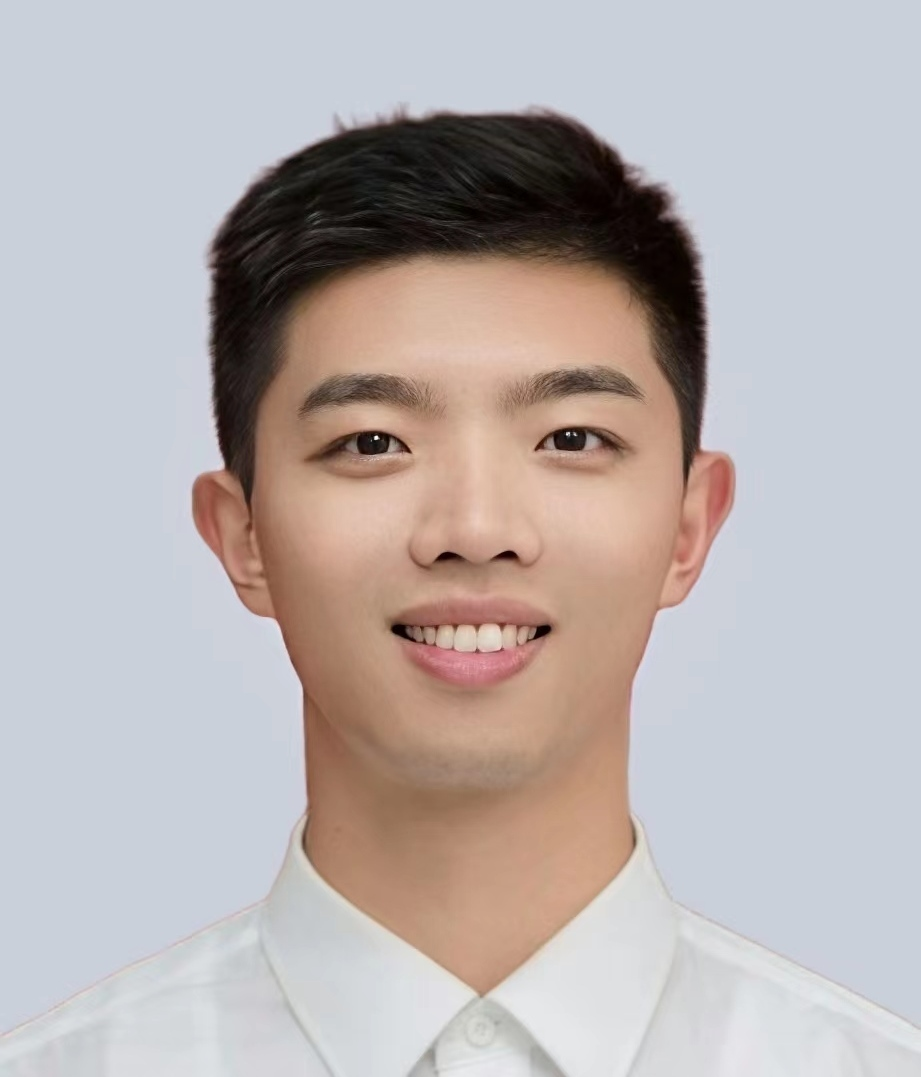}}]{Yingqian Wang}
received his B.E. degree in electrical engineering from Shandong University, Jinan, China, in 2016, the Master and the Ph.D. degrees in information and communication engineering from National University of Defense Technology (NUDT), Changsha, China, in 2018 and 2023, respectively. Dr. Wang is currently an Associate Professor with the College of Electronic Science and Technology, NUDT. His research interests focus on computational photography and low-level vision, particularly on light field image processing and image super-resolution. For more information, please visit \href{https://yingqianwang.github.io/}{https://yingqianwang.github.io/}.
\end{IEEEbiography}

\end{document}


\title{Supplemental Material of ``Dynamic High-frequency Convolution for Infrared Small Target Detection"}
\author{
    Ruojing Li, Chao Xiao, Qian Yin, Wei An, Nuo Chen, Xinyi Ying, Miao Li, Yingqian Wang
}

\maketitle

\section{Related work}

In this section, we briefly review the major studies in SIRST detection and high-frequency operators.

\subsection{Single-frame Infrared Small Target Detection}
SIRST detection methods include traditional methods and DL-based methods \cite{wang2019miss_MDvsFA, zhang2022isnet}. Traditional methods mainly include filter-based \cite{Rivest1996Detection_Top-hat, deshpande1999max, zhu2020balanced}, local saliency-based \cite{2015Infrared_PLCM, han2020infrared, yang2023small}, and sparse low-rank decomposition-based \cite{2013Infrared_IPI, 2018Infrared_NRAM, zhao2021three} approaches. They achieve good detection performance when facing specific scenarios by pre-modeling the targets and backgrounds in them, but suffer performance degradation on diverse backgrounds and targets. DL-based methods have significant performance superiority due to their powerful feature learning capabilities. There are various structures of SIRST detection networks, which are mainly divided into three types, including CNN-based methods, CNN-Transformer hybrid (CNN-T) methods, and Mamba-based methods. DNANet \cite{li2022dense} is designed with a dense nested interactive module and a channel-spatial attention module to exploit contextual information. ILNet \cite{li2025ilnet} is introduced with an interactive polarized orthogonal fusion module to integrate more low-level features into deep-level features. MTU-Net \cite{wu2023mtu} is proposed as a multilevel TransUNet for multilevel feature extraction. APTNet \cite{zhang2025aptnet} is an adaptive partial Transformer network based on the U-Net architecture for contextual information integration. Mamba-in-Mamba \cite{chen2024mim} is customized with a nested structure, using outer and inner Mamba blocks to capture global and local features. IRMamba \cite{zhang2025irmamba} is an encoder-decoder architecture featuring pixel difference Mamba and a layer restoration module for SIRST detection. These methods fit the data based on machine learning, yet ignore the explicit discriminative modeling of HFCs, which may lead to insufficient robustness of the networks.

\subsection{High-frequency Operators}\label{operators_Review}
High-frequency operators record the modeling of key information and usually use image gradients or derivatives to extract HFCs from images. They play an important role in edge detection \cite{canny1986computational}, image sharpening \cite{sobel19683x3}, object detection \cite{ferrari2008groups}, and image segmentation \cite{bertasius2016semantic}. Early methods in low-level vision tasks use the first- or second-order derivatives, e.g., Sobel \cite{sobel19683x3} and Canny \cite{canny1986computational}, as the operators to perceive grayscale variations in certain directions. Early high-level vision methods use morphological filtering operators, e.g., Top-hat \cite{Rivest1996Detection_Top-hat} and Laplacian of Gaussian, to capture the interested HFCs. Both are constrained by hand-crafted parametric rigidity. Recently, with the rapid development of deep learning, few work has explored the combination of traditional high-frequency operators and neural network layers. In edge detection, pixel difference convolution (PDC) \cite{su2021pixel} integrates fixed pixel difference operations into convolution operations for enhanced performance. In SIRST detection, ALCNet \cite{dai2021attentional} introduces conventional local contrast measurement with fixed operations into the network. IRMamba \cite{zhang2025irmamba} reuses the fixed pixel difference operation in the Mamba structure for SIRST detection. Although these methods improve the performance of the corresponding networks in different tasks, they remain constrained in adapting to various scenes and targets, since the high-frequency operators used in them still have fixed parameters like the operators in traditional methods.

\section{Methodology: Supplementary Details}

\begin{figure*}[t]
\centering
\includegraphics[width=1\textwidth]{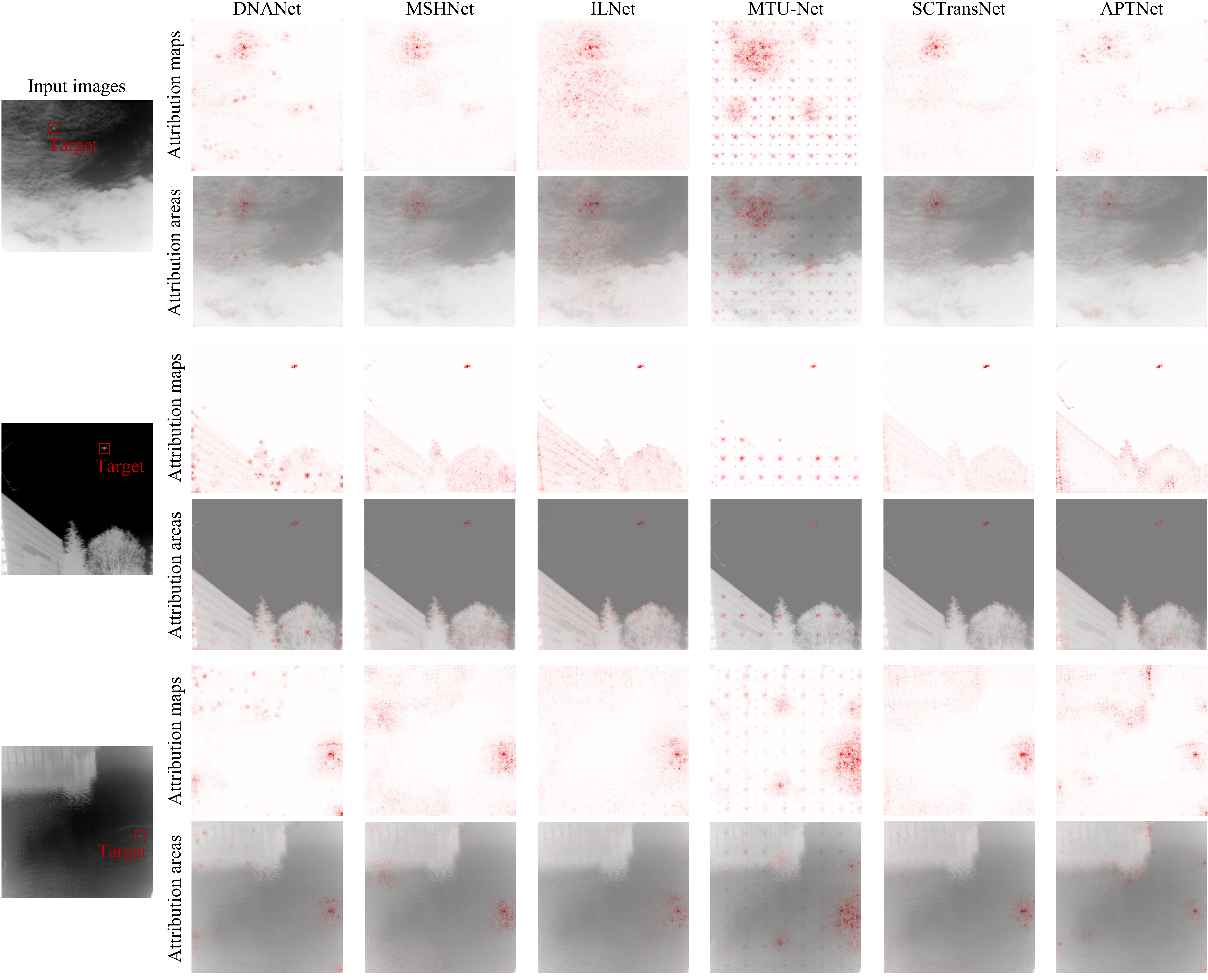}
\caption{Attribution visualizations of different methods for the predictions of target regions in different images.}
\label{attribution_supp}
\end{figure*}

\subsection{Attribution Analysis}
We conducted attribution analysis on various networks, including DNANet \cite{li2022dense}, MSHNet \cite{liu2024infraredMSHNet}, ILNet \cite{li2025ilnet}, MTU-Net \cite{wu2023mtu}, SCTransNet \cite{yuan2024sctransnet}, and APTNet \cite{zhang2025aptnet}. The attribution maps are shown in Fig. \ref{attribution_supp}. The observations are consistent with those in the main text, demonstrating that the intensity gradient regions (high-frequency regions) are worth focusing on and learning. That is, it is important for SIRST detection networks to learn representations of HFCs.

\subsection{Drop-in Replacement of DHiF}
DHiF functions as a drop-in replacement for standard convolution layers in SIRST detection networks. Then, which convolutions are optimal for replacement, and how to use DHiF? To address the two questions, it should be clear that DHiF captures information from local details (e.g., grayscale variations). In SIRST detection networks, shallow layers predominantly capture fine-grained texture variations, while deeper layers specialize in high-level semantic abstraction. Consequently, it is better to replace convolutions in encoders with our DHiF to capture the HFS, since the details of targets tend to disappear in high-level features.

\begin{figure}[t]
\centering
\includegraphics[width=0.73\columnwidth]{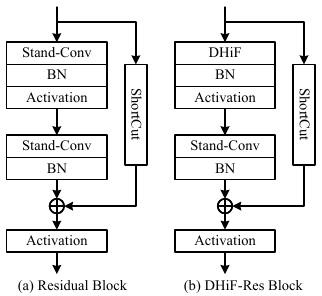}
\caption{Structures of (a) residual block and (b) DHiF-Res block.}
\label{DHiF_Res}
\end{figure}

Furthermore, the residual block \cite{he2016deep} is the important component of most SIRST detection networks. To fully leverage DHiF, it can be embedded into the residual block by replacing the first standard convolution, and form a new dynamic high-frequency residual block (DHiF-Res block). The structures of the residual block and our DHiF-Res block are shown in Fig. \ref{DHiF_Res}. DHiF-Res block can be defined as

\begin{equation}
    \label{equ_alfres}
    Res_{DHiF}(\boldsymbol{X}) = \sigma\bigg(\kappa(\boldsymbol{X}) + \gamma\Big(
    \sigma\big(DHiF(\boldsymbol{X})\big)\Big)\bigg),
\end{equation}
where $DHiF(\cdot)$ denotes the cascaded composition of DHiF and Batch Normalization (BN) layer, and $\gamma(\cdot)$ denotes the cascaded composition of standard convolution and BN layer. $\sigma$ and $\kappa$ represent the activation function and the shortcut connection, respectively.

In this way, SIRST detection networks are improved by enhancing the discriminative representation learning of different HFCs.

%

\begin{figure*}[t]
\centering
\includegraphics[width=1\textwidth]{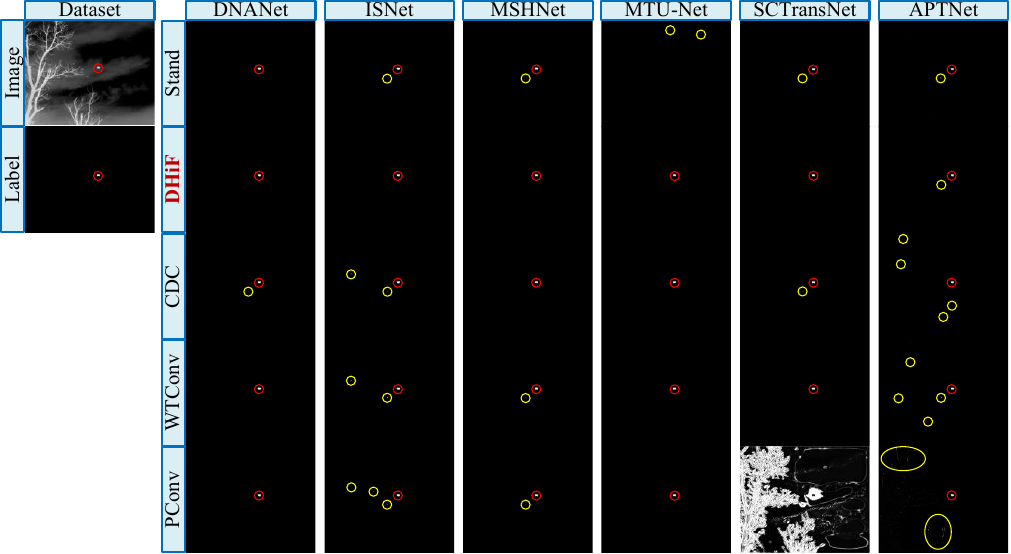}
\caption{Visual comparison of different methods with different convolutions in scene 1. The red circles mark the targets, and the yellow circles mark the false alarms. More comparison results in different scenarios are shown in the supplementary material.}
\label{vis1}
\end{figure*}

\begin{figure*}[t]
\centering
\includegraphics[width=1\textwidth]{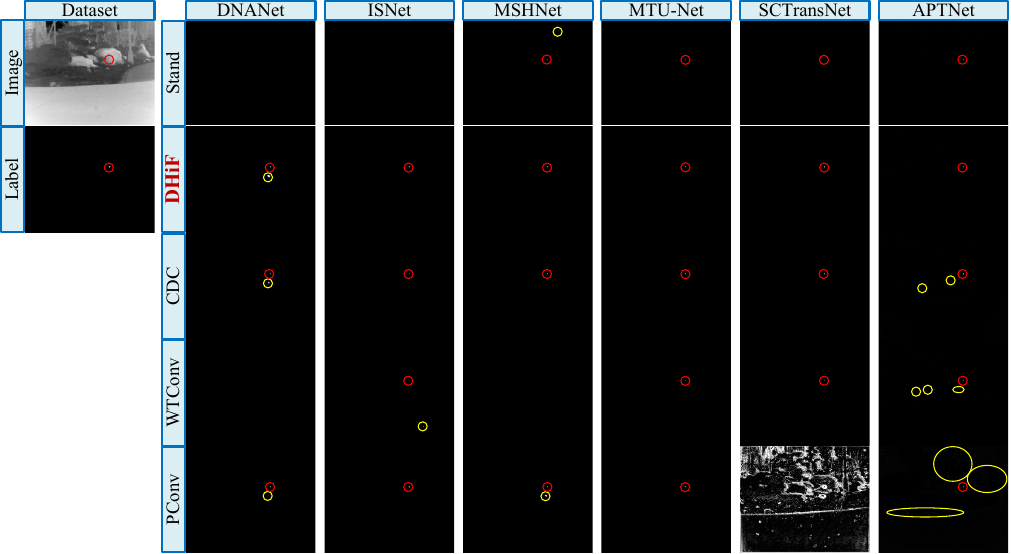}
\caption{Visual comparison of different methods with different convolutions in scene 2. The red circles mark the targets, and the yellow circles mark the false alarms.}
\label{vis2}
\end{figure*}

\section{Additional Experiments}
\subsection{Metrics in Experiments}

$IoU$ and $nIoU$ evaluate the profile description ability of the algorithm, which are defined as
\begin{equation}
    \label{equ_IoU}
    IoU =\frac{A_i}{A_u}, nIoU=\frac{1}{N}\sum_{j=1}^{N}\frac{a_i^j}{a_u^j},
\end{equation}
where $A_i$ and $A_u$ denote the interaction areas and the union areas in all test samples, respectively. $a_i^j$ and $a_u^j$ represent the interaction areas and the union areas for the $j^{th}$ sample.

$P_d$ and $F_a$ measure the abilities of algorithms to distinguish between true and false targets, which are defined as
\begin{equation}
    \label{equ_pdfa}
    P_d = \frac{T_{TP}}{T_{All}}, F_a = \frac{P_{FP}}{P_{All}},
\end{equation}
where $T_{TP}$ and $T_{All}$ are the true positive predicted target number and the total target number in the test set. $P_{FP}$ and $P_{All}$ are the false positive predicted pixel number and the total pixel number in all test samples.

\subsection{Implementation Details}
We implemented all methods in PyTorch on a computer with an Intel Xeon Gold 6328H CPU @ 2.80GHz and two Tesla V100s PCIe 32GB GPUs. In detail, the networks were trained for 500 epochs with a batch size of 16. The input image was cropped to $256\times 256$ resolution during training. We used the Adam optimizer \cite{bearman2016s} with an initial learning rate of $5\times10^{-4}$. The learning rate decayed after every 50 epochs with a decaying rate of 0.5. We initialized the weights of the convolution layers using the Kaiming method \cite{he2015delving}.

The IRSDT-1k \cite{zhang2022isnet} dataset includes 1001 real infrared images (800 for training and 201 for testing) with a resolution of $512\times 512$, and the NUAA-SIRST \cite{dai2021asymmetric} dataset includes 427 real infrared images (213 for training and 214 for testing) of varying resolutions.

\subsection{Qualitative Results}
The qualitative results are shown in Figs. \ref{vis1} and \ref{vis2}. Compared with networks with only standard convolutions, those with DHiFs perform better in accurately predicting targets. Compared with networks with other specially designed convolutions (i.e., CDC \cite{zhang2024mdigcnet}, WTConv \cite{finder2024wavelet}, and PConv \cite{yang2025pinwheel}) for SIRST detection, those with DHiFs perform better in suppressing false alarms. Therefore, our DHiF achieves superior detection performance in various networks.

\subsection{Ablation Study}\label{Additional_ablationStudy}
In this part, we analyzed several import issues about our DHiF, including the DHiF-Res block structure, the kernel size of DHiF, the nonlinear mapping of DHiF, and the model complexity and runtime. To ensure reliable comparisons, all ablation studies are conducted uniformly on the typical baseline model, i.e, DNANet \cite{li2022dense}. The modified DNANet structure involves the residual blocks in the second and third layers of its encoder.

\subsubsection{DHiF-Res block structure}
To investigate the optimal structure of the DHiF-Res block, we altered the position of DHiF within the block and compared the detection performance of different variants, with results shown in Table \ref{tab:block}.

\begin{table}[t]
\caption{Detection results achieved by variants of DHiF-Res block with different structures on the NUAA-SIRST dataset.}
\label{tab:block}
\centering
\renewcommand{\arraystretch}{1.1}
\begin{tabular}{c c c c c}
\hline
structure & $IoU$ & $nIoU$ & $P_d$ & $F_a$ \\
\hline
DHiF + standard & 77.68 & 79.03 & 95.42 & 1.53 \\
standard + DHiF & 74.55 & 77.97 & 93.13 & 1.38 \\
\hline
\end{tabular}
\end{table}

As indicated, implementing DHiF first followed by standard convolution yields significantly superior performance in terms of $IoU$, $nIoU$, and $P_d$. Performing standard convolution after DHiF facilitates further acquisition of discriminative representations of HFCs from locally differentiated high-frequency information extracted by DHiF.

\subsubsection{Kernel size of DHiF}
To explore the impact of the convolution kernel size on DHiF performance, we tested different kernel sizes (e.g., 3, 5, 7) and compared the performance of DHiF with that of standard convolution. The corresponding results are presented in Table \ref{tab:ks}.

\begin{table}[t]
\caption{Detection results achieved by variants of DHiF with different kernel sizes on the NUAA-SIRST dataset. The best results are in \textbf{bold}.}
\label{tab:ks}
\centering
\renewcommand{\arraystretch}{1.1}
\setlength{\tabcolsep}{6pt}{
\begin{tabular}{c c c c c c}
\hline
Kernel size & Convs & $IoU$ & $nIoU$ & $P_d$ & $F_a$ \\
\hline
3 & Standard & 71.97 & 78.66 & 95.04 & 6.31 \\
3 & DHiF & \textbf{77.68} & \textbf{79.03} & \textbf{95.42} & 1.53 \\
5 & Standard & 72.36 & 77.55 & 93.51 & 3.49 \\
5 & DHiF & 76.87 & 78.86 & 94.27 & 1.58 \\
7 & Standard & 74.17 & 77.26 & 93.13 & 1.26 \\
7 & DHiF & 76.80 & 79.02 & 94.27 & \textbf{1.06} \\
\hline
\end{tabular}}
\end{table}

It can be observed that DHiF delivers substantial performance gains regardless of the kernel size. However, larger kernels do not obviously improve detection performance for any type of convolution, while causing higher computational costs. This outcome relates to the specific network architecture: relying solely on convolution layers to adequately capture global information from larger receptive fields is challenging. Moreover, the operation parameter configuration of DHiF can be set consistently with that of the replaced standard convolution, requiring no special tuning.

\subsubsection{Nonlinear mapping of DHiF}
In Section II-B.2 of the main text, we infer that zero-centered filter parameters are both necessary and sufficient to make DHiF sensitive to HFCs. Here, a zero-centered distribution means that zero lies within the value interval (zero-included condition) and that values are symmetrically distributed around zero (symmetry condition). To demonstrate this point, we conducted an ablation study on the nonlinear mapping function. Specifically, we tested four variants: removing the mapping function $tanh(\cdot)$ entirely, and replacing it with GELU \cite{hendrycks2016gaussian}, LeakyReLU \cite{maas2013rectifier}, or Sigmoid functions. The corresponding results are presented in Table \ref{tab:nonlinear}.

\begin{table}[t]
\caption{Detection results achieved by variants of DHiF with different nonlinear mappings on the NUAA-SIRST dataset. The best results are in \textbf{bold}.}
\label{tab:nonlinear}
\centering
\begin{threeparttable}
\renewcommand{\arraystretch}{1.1}
\setlength{\tabcolsep}{10pt}{
\begin{tabular}{c c c c c}
\hline
Nonlinear & $IoU$ & $nIoU$ & $P_d$ & $F_a$ \\
\hline
None & 71.81 & 75.58 & 91.98 & 2.94 \\
GELU\tnote{1} & 71.12 & 77.02 & 93.75 & 3.86 \\
LeakyReLU\tnote{1} & 72.92 & 76.84 & 93.51 & \textbf{1.48} \\
Sigmoid\tnote{2}  & 72.25 & 78.28 & 93.89 & 4.13 \\
Tanh  & \textbf{77.68} & \textbf{79.03} & \textbf{95.42} & 1.53 \\
\hline
\end{tabular}}
\begin{tablenotes}
\item[1] A mapping function that dose not satisfy the symmetry condition.
\item[2] A mapping function that does not satisfy the zero-included condition.
\end{tablenotes}
\end{threeparttable}
\end{table}

The results show that only the DHiF with $tanh(\cdot)$ whose mapping range is strictly zero-centered, delivers an obvious performance gain. DHiFs with other mapping functions fail to outperform even standard convolution. This occurs because those functions do not satisfy the zero-centered distribution, and thus cannot guarantee that DHiF remains sensitive only to HFCs. As a result, they introduce interference into the feature maps, which degrades the network performance. Likewise, removing the mapping entirely causes a significant performance degradation, because the unrestricted filter parameters may assume extreme values and lose the zero-centered distribution. Therefore, functions that map filter parameters into a zero-centered interval, such as $tanh(\cdot)$, are essential for DHiF.

\subsubsection{Model complexity and runtime}
We reported the model complexity and runtime of the models with different convolutions. Specifically, we replaced one convolution in each involved residual block with different convolutions (CDC \cite{zhang2024mdigcnet}, WTConv \cite{finder2024wavelet}, PConv \cite{yang2025pinwheel}, and our DHiF), and evaluated the parameter counts (params), FPS, and runtime of each variant. The results are reported in Table \ref{tab:complexity}.

\begin{table}[t]
\caption{Model complexity and runtime comparison of different convolutions. FPS and runtime are measured based on processing $512\times 512$ images.}
\label{tab:complexity}
\centering
\begin{threeparttable}
\renewcommand{\arraystretch}{1.1}
\setlength{\tabcolsep}{8pt}{
\begin{tabular}{c c c c}
\hline
Convs\tnote{*} & \#Params & FPS & Runtime (sec) \\
\hline
Standard & 17.92 M & 8.61 & 0.1161 \\
CDC    & 17.92 M & 8.60 & 0.1163 \\
WTConv & 17.60 M & 8.57 & 0.1167 \\
PConv  & 17.78 M & 8.43 & 0.1186 \\
DHiF   & 17.93 M & 8.02 & 0.1247 \\
\hline
\end{tabular}}
\begin{tablenotes}
\item[*] A total of four convolution operations are involved in the network.
\end{tablenotes}
\end{threeparttable}
\end{table}

As shown in Table \ref{tab:complexity}, replacing the four convolutions with DHiF introduces only 0.01~MB new parameters while reducing computation efficiency by $6.85\%$. This indicates that the obvious performance improvement stems from the capability of modeling local HFCs, rather than from increasing model parameters.

\subsection{More Evidence for Effectiveness on Learning Differential Representations}
\begin{figure*}[th]
\centering
\includegraphics[width=1\textwidth]{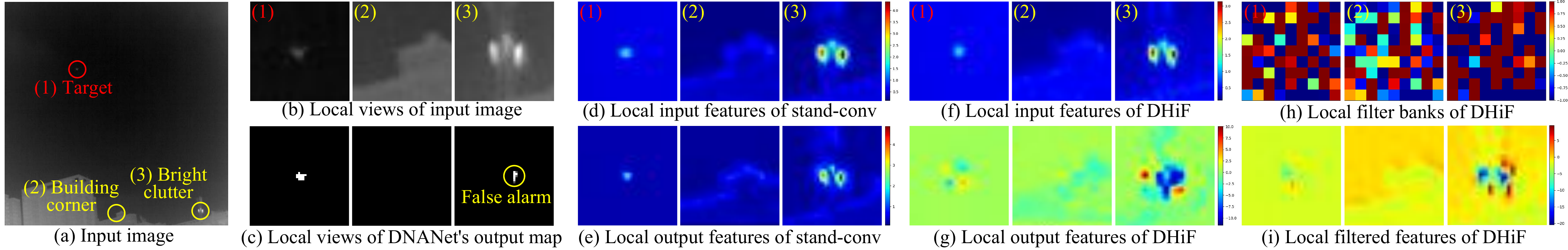}
\caption{Visualizations of features in the second levels of DNANet variants without and with DHiF. (a) The input image including (1) target, (2) building corner, and (3) bright clutter. (c) Local view of the output map generated by the original DNANet. (d), (f) Local input features and (e), (g) local output features of the first convolutions (standard convolution and DHiF). (h) Local filter banks of three HFCs are differentially sensitive to high-frequency information, since their parameter values are different and symmetrically distributed within the range of $[-1, 1]$. Therefore, DHiF has better capability in discriminatively representing different HFCs.}
\label{vis_conv}
\end{figure*}

The visualization results from the second levels of DNANet variants without and with DHiF are shown in Fig. \ref{vis_conv}, which are similar to those from the third levels shown in the main text.

\subsection{Robustness to Salt-and-Pepper Noise}
To further investigate the modeling capability of DHiF for HFCs, we replaced part of the convolutions in the DNANet encoder with CDC \cite{zhang2024mdigcnet}, WTConv \cite{finder2024wavelet}, PConv \cite{yang2025pinwheel}, and DHiF, following the same modification scheme described in Section \ref{Additional_ablationStudy}. Then, we compared their robustness under salt-and-pepper noise as shown in Table \ref{tab:salt_pepper}.

\begin{table}[t]
\caption{Detection results achieved by variations of DNANet with different convolutions under salt-and-pepper noise. The best results are in \textbf{bold}.}
\label{tab:salt_pepper}
\centering
\renewcommand{\arraystretch}{1.1}
\setlength{\tabcolsep}{10pt}{
\begin{tabular}{c c c c c}
\hline
Convs & $IoU$ & $nIoU$ & $P_d$ & $F_a$ \\
\hline
Standard & 66.48 & 70.45 & 89.69 & 6.52 \\
CDC    & 67.56 & 72.03 & 91.22 & 5.86 \\
WTConv & 69.00 & 71.12 & 88.55 & 3.00 \\
PConv  & 66.89 & 70.21 & 88.55 & 4.56 \\
DHiF   & \textbf{72.08} & \textbf{72.82} & \textbf{92.37} & \textbf{2.66} \\
\hline
\end{tabular}}
\end{table}

The comparison shows that the network with DHiF significantly outperforms all other convolutions under salt-and-pepper noise. It reduces $F_a$ by $11.33\%$ compared to the second-best result (WTConv), increases $P_d$ by $1.15\%$ over the second-best (CDC), and improves $IoU$ by $3.08\%$ relative to the second-best (WTConv). These results strongly demonstrate the outstanding capability of DHiF in modeling HFCs.

\section{Discussion}
In this section, we discuss the limitations of DHiF.

First, regarding framework compatibility, DHiF is best suited for architectures that extract features based on a large number of convolutional layers, such as CNN-based and CNN-Transformer hybrid models. DHiF can consistently improve the performance of these models. However, in frameworks that employ only few convolutions at the input layer, e.g., some Mamba-based methods \cite{chen2024mim}, DHiF may be less effective. As analysed in Section III-C.1 of the main text, applying DHiF at the input layer may disrupt the original data distribution, potentially limiting its effectiveness in such cases.

Second, as DHiF focuses on capturing local details, it faces a fundamental limitation when local clutter patterns closely resemble the target. In these situations, accurate discrimination usually requires broader contextual or semantic information from surrounding regions.
